\documentclass[journal]{IEEEtran}

\usepackage[style=ieee]{biblatex} 
\usepackage{hyperref}
\usepackage{amsmath}
 \usepackage{multirow}

\usepackage{url}
\usepackage{xcolor}
\usepackage{graphicx}  
\usepackage{caption}
\usepackage{float}  
\newcommand\tab[1][0.5cm]{\hspace*{#1}}

\usepackage{titlesec}
\titleformat{\section}
  {\vspace{1pt}\normalfont\scshape\centering}{\thesection.}{3pt}{}[\hrule\vspace{1em}]

\begin{document}

\title{Convolutional Neural Networks Towards Arduino Navigation of Indoor Environments \\ }

\author{Michael Muratov,
        Sachkiran Kaur,
        Michael Szpakowicz\\
        Supervised by Professor Arnold Rosenbloom\\Computer Science Department\\University of Toronto\\ Toronto, Canada}

\markboth{\MakeSentenceCase{{CSC}498 {R}eport, {U}niversity of {T}oronto, {A}pril 2019}}
{Shell \MakeLowercase{\textit{et al.}}: Bare Demo of IEEEtran.cls for IEEE Journals}

\maketitle

\begin{abstract}In this paper we propose a number of tested ways in which a low-budget demo car could be made to navigate an indoor environment. Canny Edge Detection, Supervised Floor Detection and Imitation Learning were used separately and are contrasted in their effectiveness. The equipment used in this paper approximated an autonomous robot configured to work with a mobile device for image processing. 
This paper does not provide definitive solutions and simply illustrates the approaches taken to successfully achieve autonomous navigation of indoor environments. The successes and failures of all approaches were recorded and elaborated on to give the reader an insight into the construction of such an autonomous robot. 
\end{abstract}


\section{Introduction}

\IEEEPARstart{S}{elf} driving cars have become a popular topic of discussion in recent years due to their high accuracy, and the possibility of complete automation they bring to public transportation.
Modern self-driving cars process thousands of gigabytes of data every day from an array of sensors, such as high quality cameras, GPS, LIDAR, and RADAR, to map out and navigate though a three-dimensional environment.
The full spectrum of data is passed through machine learning algorithms which give directions to the car, often more accurately than a human could. 
This is due to the system's more accurate perception of the road, and its super-human reaction speeds.
These multiple sources of data make the self driving car incredibly aware of its surroundings, but this comes with high purchase and maintenance costs.
Unlike modern autonomous vehicles, the resources used in this project primarily include a re-purposed RC car, an Arduino UNO microcontroller, a motor-driving shield, a Bluetooth module, and a smartphone with a singular rear facing camera.

In this paper, we outlined our method for creating a self driving car which is able to navigate corridors and stay on track with where it is heading. In most cases, the car is able to perform at a level similar to that of a human driver, and it has proven to be effective in new environments and lighting conditions.

For the duration of our project, we have not followed any past research papers on the subject of autonomous driving, and thus, this paper will outline the entire development process of such a self driving vehicle.
The source code for all aspects of the project is available online, and will be referenced at the end of the report for anyone looking to extend on our work.

\section{Getting things moving}

The mechanics and limitations of the car where an immediate hurdle in terms of the repeatability of our experiments.
The turning system on the car did not allow reliably tight or accurate turns, which made controlling the car difficult even to a human driver. This was partially mitigated by incorporating additional batteries, beyond what was intended by the manufacturer. \\

\begin{figure}[H]
\centering
\includegraphics[width=0.45\textwidth]{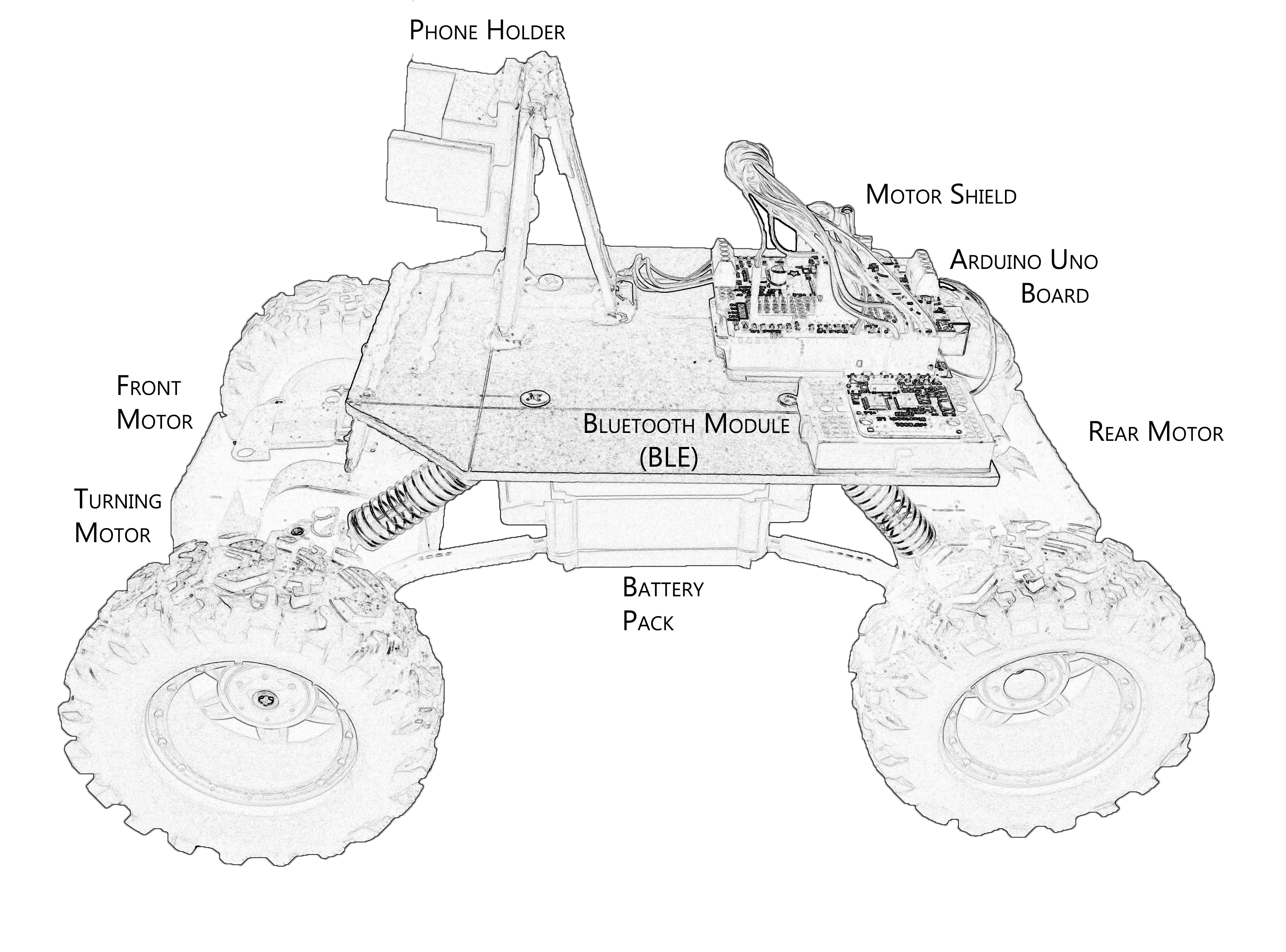}
\caption{Car components}
\end{figure}

The car itself was controlled by an Arduino UNO board through an Adafruit Motor Shield connected to the car's various motors.
The car had two DC motors, one for turning each pair of wheels, along with a DC motor mounted to the steering bracket on the front axle.
The Arduino board processed commands sent in from an Android phone via the Adafruit Bluetooth Low Energy (BLE) module.
This setup allowed the car to either be controlled by a human through touch-screen controls, or to be controlled autonomously by parsing camera input data.
The ability to control the car remotely proved to be essential in providing large amounts of training data quickly for the car's neural network.

The aforementioned components were secured onto the upper portion of the car chassis, along with a safety switch which could immediately cut the power across the components (\textit{Fig. 1}).
The weight of the components would cause the RC car's suspension mechanism to sag, which affected the car's ability to turn and drive straight.
The problem was rectified by adding support beans around the suspension, effectively disabling it, and ensuring the upper chassis remained parallel to the ground.
The task would have been more difficult if the system needed to account for the car's pitch and roll.
This setup was deemed acceptable in approximating a real car's functions while ensuring minimal development expenses.

\section{Computer Vision}

\begin{figure}[H]
\centering
\includegraphics[width=0.4 \textwidth]{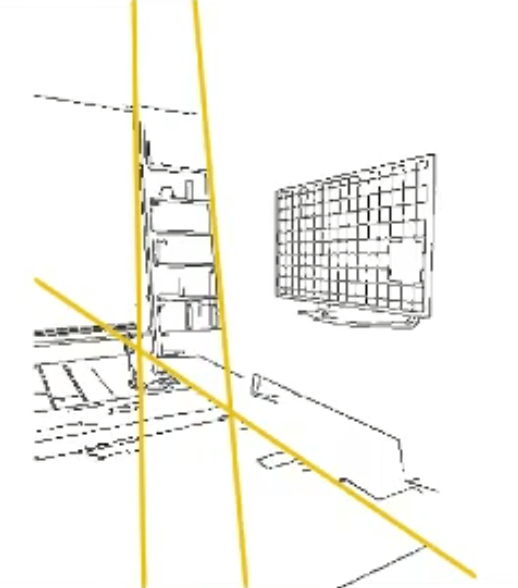}
\caption{Our Canny Edge Detection using OpenCV}
\end{figure}

In this section we outlined the methods we explored in our search for a reliable algorithm for navigating indoor environments.
The first approach we attempted to implement was the traditional, computer-vision-aided feature extraction with Canny Edge Detection [1] and Blob Detection for identifying individual components of a scene.
This approach would allow for the segmentation of an image into essential components such as walls and floors and allow for direct hard coding of car behavior in key situations. 

Although this approach has been shown to work in practice, this method assumes that lighting conditions and environment geometry stay relatively constant.
A similar paper released in April 2018, ``Image-Based Segmentation of Indoor Corridor Floors for a Mobile Robot'' [2], achieved highly accurate results in drawing a separation line between floor and wall segments of an image.
The paper outlined a unified algorithm utilizing Image Thresholding, Canny Edge Detection and Graph Based Segmentation.
The paper listed some disadvantages, namely poor performance on highly textured floors, which would require additional considerations when implementing it for actual robot use.
Above is an image of a scene and the lines generated by Canny Edge, providing a considerably accurate description of the scene (\textit{Fig. 2}).

The limitation of this process lies in keeping track of which lines stay consistent from image to image.
The reading for one image using Canny Edge Detection [1] may be fairly accurate, but the detected lines may change even between  images.
The conclusion reached was that this would be an unstable and, ultimately, time-consuming approach to represent the scene. It was also important to not simply replicate the procedure of the recent April 2018 paper [2], even if our implementation would result in higher performance than suggested by the authors.

\section{Convolution Neural Net Floor Detection}

The detection of the floor in a general environment required a more advanced approach than we first realized.
This led us to leveraging the industry-grade neural networks which have been shown to handle a variety of tasks with near human accuracy.
This approach requires a large amount of training data to optimize the neural network's internal weights.
In this case, the data would be in the form of images of what is ahead of the car, along with some human generated labels that would allow the algorithm to differentiate between the floor and walls in the scene.

In order to obtain images of the scene, a mobile phone was strapped to the front of the car and set to capture pictures at a constant rate.
The human driver controlled the car remotely, navigating the car around the facility while it captured training images roughly every four seconds.
This approach would produce approximately 150 training images per ten minutes of recording.
In addition to capturing raw images, a human would be required to label said images by creating polygonal masks with red polygons covering the floor area and blue polygons covering the walls (\textit{Fig. 3}). 

\begin{figure}[H]
\centering
\includegraphics[width=0.25\textwidth]{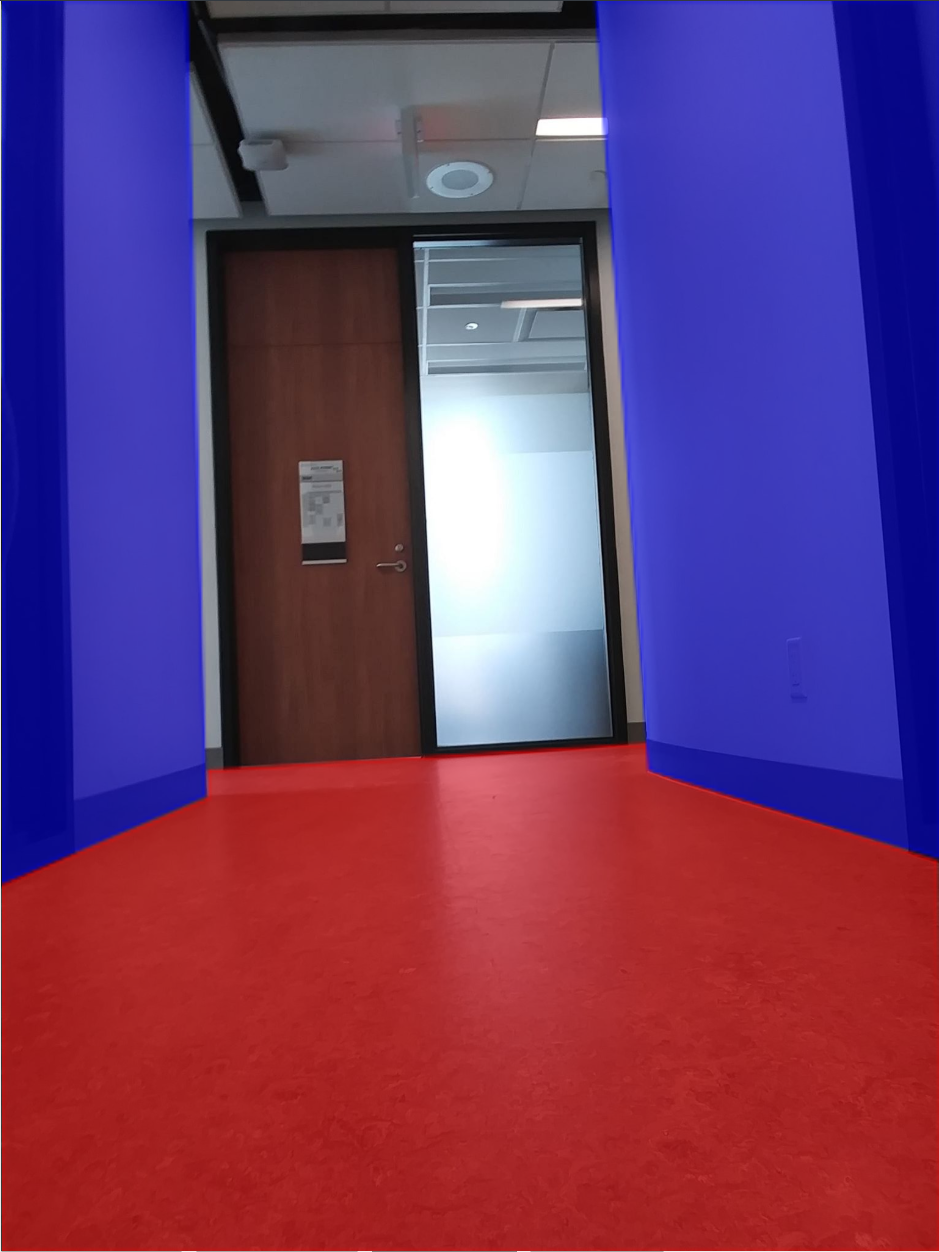}
\caption{Method for labeling of floor and walls in images}
\end{figure}

This approach proved to be extremely labor intensive as for each image, a human classifier had to create multiple labeling masks.
With the amount of data that a Convolutional Neural Network (CNN) requires to learn the relationship between pixels in the scene, this approach would require manually classifying thousands of images with no guarantee of conclusive results.
The results of the CNN floor detection shown bellow were obtained from an algorithm which learned from only around forty classified training images. 

\begin{figure}[H]
\centering
\includegraphics[width=0.45\textwidth]{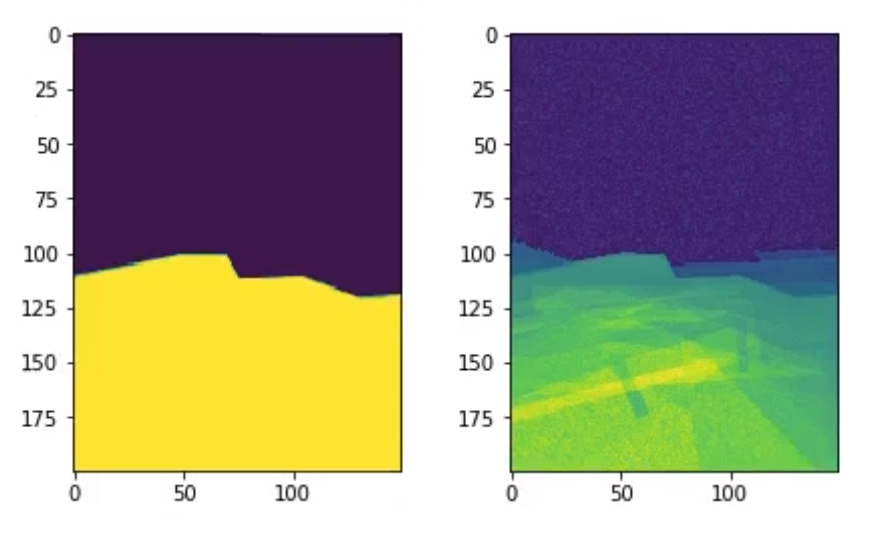}
\caption{Convolutional Neural Net Results}
\end{figure}

The top image is a correct floor mask given the input image, and the bottom is the output of our CNN.
Given that the network was overtrained on a small set of images it is not surprising that it was fairly confident in most of the image, especially of the segment in the lower half where it is most common to encounter the floor.
As stated above, it would have been a gamble to use our time for labeling thousands of pictures to get a better idea if this algorithm was a viable solution to our problem.

As an aside, the benefits of this kind of approach include recognition of complex environment settings as well as simple scenes with very little details.
Given enough diverse labeled training images, this approach would, in theory, produce fairly accurate driving commands, due to its increased awareness of objects in the environment.
This method alone, however, would have trouble detecting and turning into corners; the placement of walls may give little to no indication of a possible corner, forcing the car to instead rely on lighting and other cues to decide on taking a turn.

\section{Imitation Learning}

\subsection{Collecting the Data}

Learning from our past approach, our next attempt had to minimize the time spent on acquiring training data.
Anything as complex as labeling the environment manually could not be used, as it would make the development time go over our allotted limit.
This made us consider having the car learn to navigate itself not by separating the objects in the scene into their own classes, but by imitating how a human driver would control the car in a given scene.
This would require two sets of data: images that the car would see, and the corresponding actions that a human driver would give the car in its scene.
The most difficult part of this approach was the synchronization between these two sets of data.
Our setup consisted of two mobile devices -- one on the car capturing training images, and a second phone used to control the car manually via an on-screen joystick.

In order to synchronize the results from two mobile devices, a central server was created as a middle ground.
The images from the first phone and the commands from the second phone were paired together on this server via their timestamps. The server formatted the data accordingly into a labelled set, ready for training.

The image-taking application from our second approach was improved to instead record video, which led to a massive increase in data collection speeds.
This change allowed the collection of approximately 18000 images in 10 minutes, which is over 100 times more than the previous method.
Entire video files were uploaded to the server where a slicing function divided the video into the required amount of images.
It was decided to extract images at quarter-second intervals to ensure the images were different enough from each other and could therefore represented unique data points.
A timestamp for each image was obtained through adding the timestamp of video to the time after which the image was recorded in the video.
The image and its corresponding timestamp were then saved in a database on the server.
The next step was to label the images at each timestamp.

The controlling phone would send command packets to the car's BLE module containing the speeds and directions for each of the three motors.
Each command was timestamped and saved in batches on the phone.
A whole batch would be periodically sent from the phone to the server, which would then enter each command and its timestamp into the database.
This labeled data originated from the controller app's joystick positions.
From this data, 9 sectors were drawn as distinct actions the car would be able to emulate.
Below is a chart of what the data points and sectors looked like when viewed by the server.

\begin{figure}[H]
\centering
\includegraphics[width=0.35\textwidth]{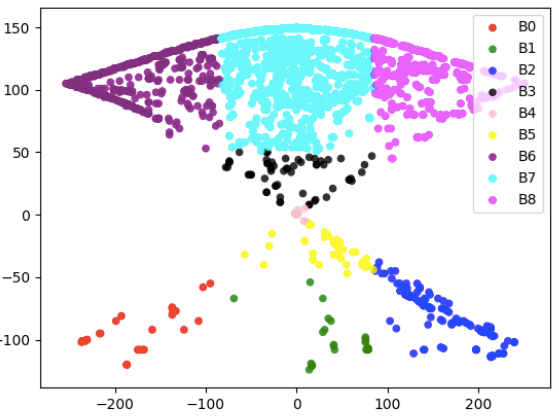}
\caption{Results received from the controller phone by the server}
\end{figure}

The server would transform the position on the joystick to one of these 9 sectors bellow.

\begin{center}
\small
\begin{tabular}{lll}
0) Backwards Left & 1) Backwards & 2) Backwards Right\\
3) Slightly Forwards & 4) Stop & 5) Slightly Backwards \\
6) Forwards Left & 7) Forwards & 8) Forwards Right
\end{tabular}
\end{center} 

For each image in the database a point with the closest timestamp was matched to it as the corresponding label.
If no point was sufficiently recent, the image was not used for the training set.
This approach was very effective, as it was able to cut out all the sections of video when the car was not actually being controlled, saving us time by not having to manually cut the video.
By the end of this process, each image was labeled with one of the 9 actions mentioned above.
A folder of timestamped images along with a file of timestamped labels could be extracted from the server at any time.
Each timestamped image corresponded to one timestamped label and vice versa.
The information contained in the label for each image is an approximation to what a human would choose to do given the image, taken while the car was driven via the joystick application.

As an aside, the improvement in how much training data we could collect still wouldn't have helped us in our previous approach, since a human would have still been required to label the images.
With this approach, we had label data automatically through simply manually driving the car.

\subsection{Convolutional Neural Network}

Given the data collected from manually driving the car, we were now able to use a CNN on the set of input image data and compare it to the corresponding list of labeled directions.
This supervised learning approach attempts to replicate the results produced by the human driver.
This made the testing data was imperfect, but for the purposes of our project, near human performance was acceptable.

The algorithm consisted of a Keras neural network which took as input a 1D array of pixels from a single grayscale image and produced 9 confidence values, one for each of the motions the car could produce.

\begin{figure}[H]
\centering
\includegraphics[width=0.50\textwidth]{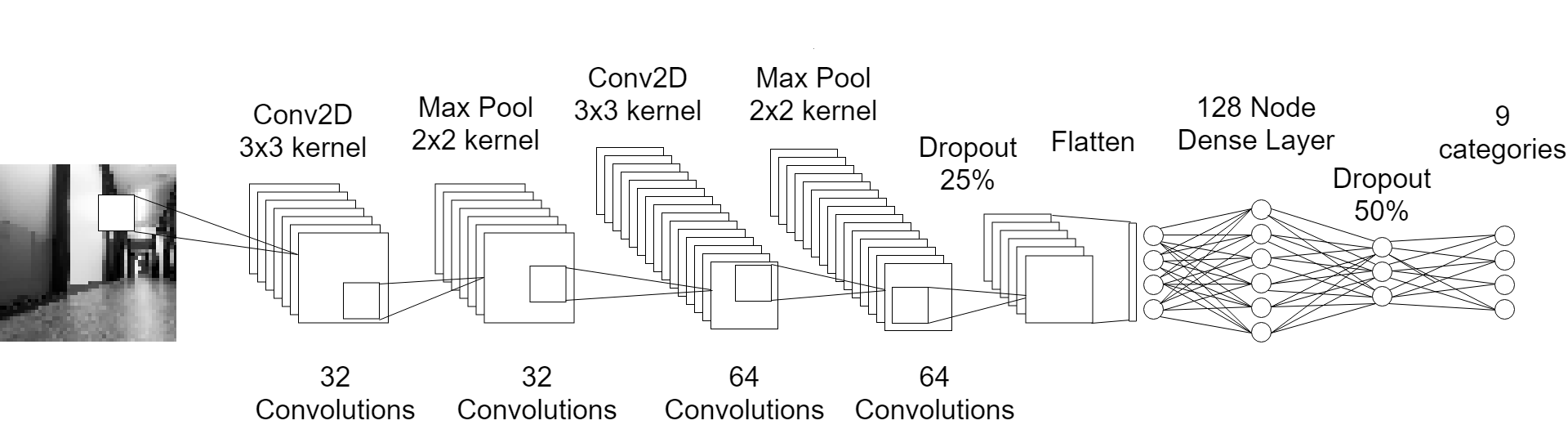}
\caption{Convolutional Neural Network model visualization}
\end{figure}

The model made use of two convolution layers separated by max pooling layers which led the algorithm to focus on potentially important features of the image.
The results of each layer being applied to an image can be seen below.

\begin{figure}[H]
\centering
\includegraphics[width=0.40\textwidth]{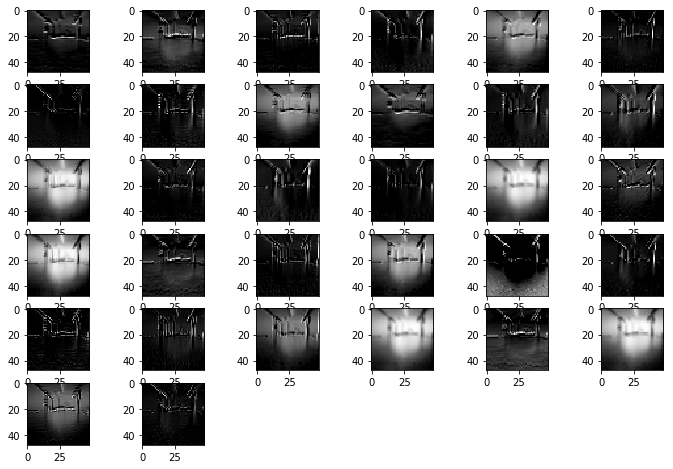}
\caption{First Convolution Layer applied to one image}
\end{figure}

\begin{figure}[H]
\centering
\includegraphics[width=0.40\textwidth]{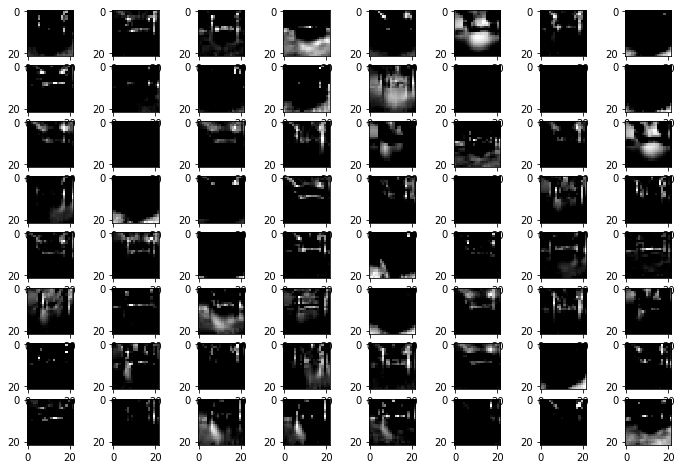}
\caption{Second Convolution Layer applied to one image}
\end{figure}

\tab The result of the first convolution layer (Fig. 7) shows a blurring of the initial image and the detection of edges in some instances.
The result of the second layer of convolutions (Fig. 8) seemed to focus on blotches of color that were of importance in the image.
Testing with additional convolution layers yielded a decrease in the algorithm's performance.
The convolutions were subsequently flattened and converted into a deep neural network with a dense hidden layer of 128 nodes.
A dropout layer was used to improve generalization of the network by only relying on half of the information produced by the hidden layer to make an educated guess.
The dropout later was then densely connected to a layer of 9 nodes, representing the 9 movements of the car.
The network was run with the Adam optimizer and a sparse, categorical cross-entropy loss function on up 100 epochs.
The resulting accuracy on a 33\% cross-validated test set was 95\%.

Now that the python model was trained, the next step was to transfer it to the Android phone used to capture images. Using the Tensorflow Lite API, it was a simple process to load a model.tflite file onto the memory of the phone.
This model would run in an application using OpenCV to obtain new images from the rear camera at a rate of multiple times a second.
The result was the phone was able to give consistent results at superhuman speed. 

Testing the model on the car revealed that the neural network performed well at avoiding crashes, and was well equipped to traverse different types of hallways with differing features and lighting.
The model was less equipped to navigating open spaces, but it would eventually find its way onward while avoiding making contact with any objects it encountered.

\subsection{Improvements to data processing}

Due to a limited data set for actions excluding forward motion, data was duplicated to establish equal representation of all actions.
This allowed for less frequent, but equally important actions to remain relevant in the network's decision making.
Subsequently, this caused a increase in accuracy ranging from 2-5\%, allowing us to reach up to 95\% test set accuracy. 

\begin{figure}[H]
\includegraphics[width=0.49\textwidth]{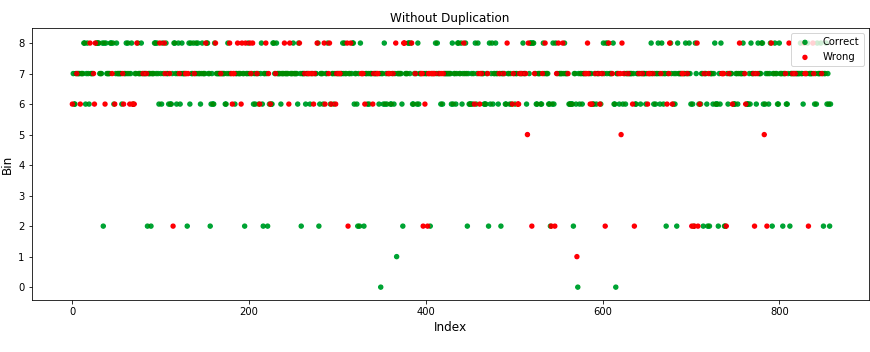}\\
\caption{Results of the test set without duplication}
Many actions are not well represented and therefore are predicted to be other more popular actions rather than themselves
\end{figure}

\begin{figure}[H]
\includegraphics[width=0.49\textwidth]{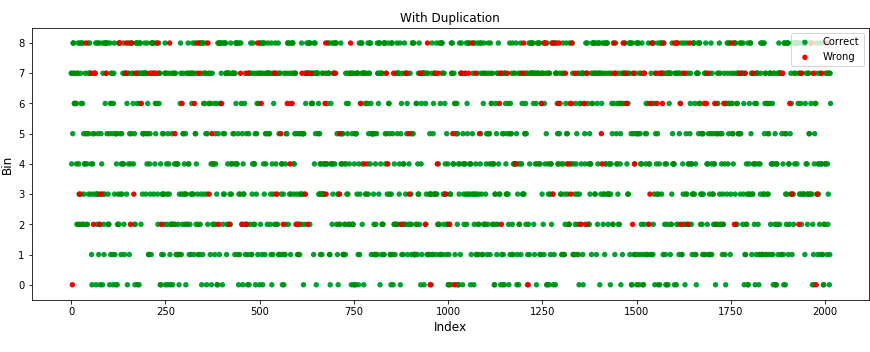} \\
\caption{Results of the test set with duplication}
The duplication of training data of less frequent results increased accuracy due to less likely occurrences being considered for classification more frequently.
\end{figure}

Another improvement to the generalization of this approach involved normalizing the input images using Histogram Equalization.
This allowed for the navigation of environments of different lighting without suffering a decrease in performance. 

\begin{figure}[H]
\centering
\includegraphics[width=0.4\textwidth]{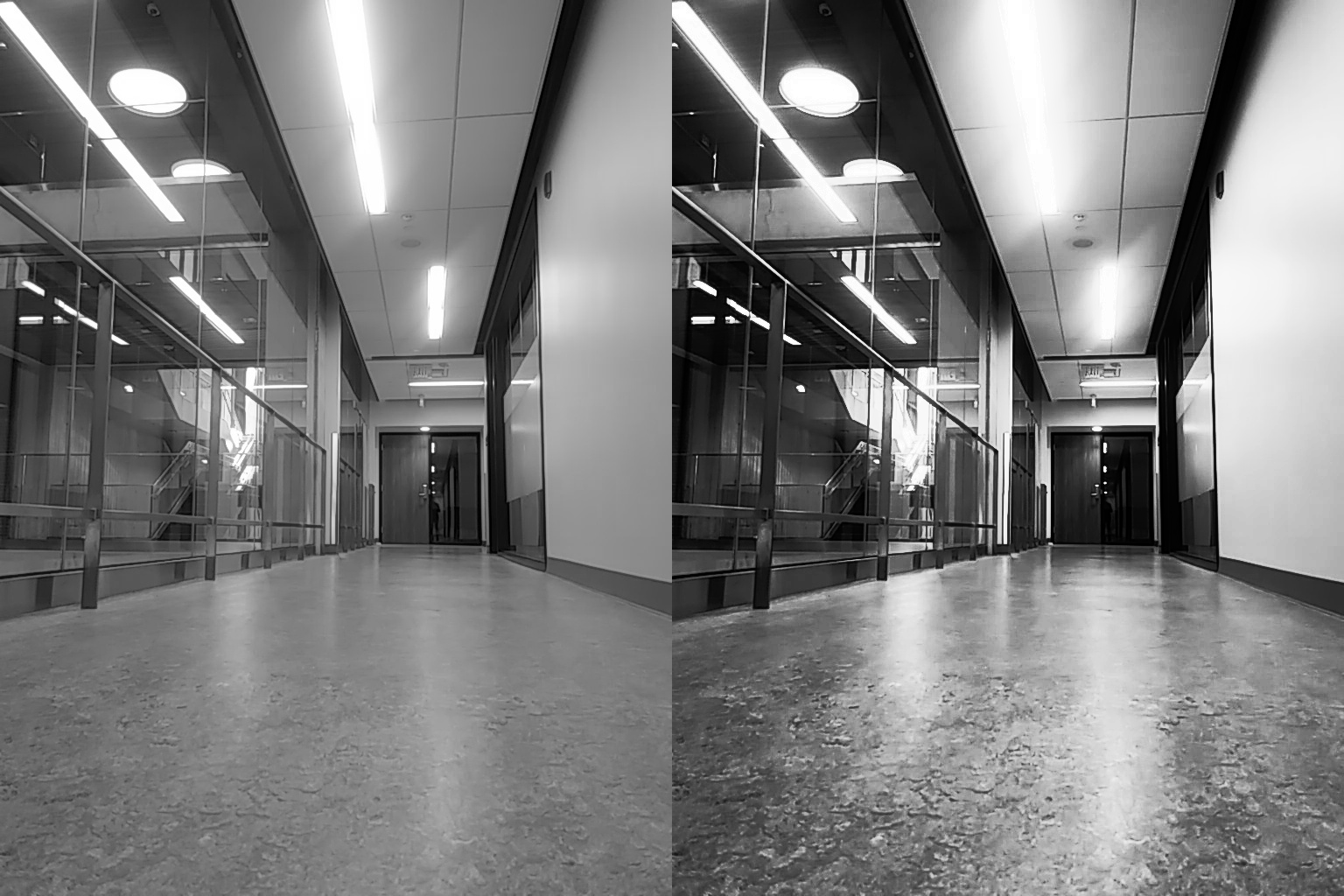}
\caption{Before and after applying histogram equalization.}
\end{figure}

As seen in the two images, more contrast differentiation is exposed which aids the CNN to pick up on patterns more easily and to treat environments of different lighting with the same level of confidence.
This change resulted in another minor increase in test accuracy.

In addition to the duplication of data, some images and corresponding labels were flipped along the y-axis to create a new set of labeled data never seen by the network.
This, however, ended up decreasing the accuracy of the model, so the method was abandoned.

\subsection{Additions to the Algorithm}

This section serves to outline our failed attempts at improving the imitation learning algorithm past its capabilities in the previous section.

\subsubsection{Multiple Past Image Input}

The CNN allows for inputs with multiple channels, such as RGB images.
We did not need to use these channels since the images we use are grayscale, but we could use them to provide the CNN with multiple images, each representing the current or a previous point in time.
In our experiments, we grouped 10 images in a chronologically-ordered sequence.
The whole sequence spanning 2.5 seconds in total, and it was assigned the label of the last image to the 
sequence.

In theory, this would allow our algorithm to gain a short term memory, and make decisions not only on the current image and its relationship to the label, but also upon the previous images as reference.
This approach showed a huge decrease in performance in comparison to the single image approach.
It is possible that this method could be implemented differently to work better than the single channel input, but our results have shown that this approach did not provide the improvement we anticipated.
The images below show the first, middle, and last indices of a set of 10 images and how they change as you look further into the set. 

\begin{figure}[H]
\centering
\includegraphics[width=0.40\textwidth]{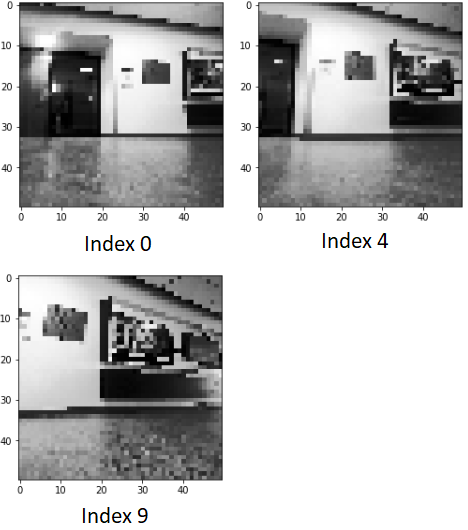}
\end{figure}

\subsubsection{Canny Edge Detection}

Applying a Canny edge generated image directly to the algorithm produces very poor results due to sever loss of information. An approach we did not have time to fully implement was supplying both the original, and canny edge generated images as two channels for the CNN to process as input.

In theory, this would allow the neural network to learn edges quicker and with more accuracy.
We did not have time to implement this approach due to the surprising failure of attempting the previous approach, using a set of past images as input.
Due to those failed attempts, the implementation of the canny approach was delayed.

\subsubsection{Additional Sensors}

Having built a version of the car navigating solely with ultrasonic sensors, we attempted to enhance the neural network aspect of the car using distance input from the sensors.
The easiest way to handle this was to simply override the neural network.
Whenever the sensors detected that the car was too close to an object, the CNN output was ignored, and the car backed away from the direction of the object.
This approach was not effective and would cause the car to perform repetitive motions, as it struggled to manage the control from the sensors and neural network. 

The other approach was to train two neural networks, one on the images and another on sensor data and to merge the results as the result of the algorithm.
During the training of the car, we realized that the data from the sensors was extremely unreliable.
In addition, this approach would force us to find an appropriate ratio between the importance of the image data versus the importance of sensor data.
In the end, this approach was not fully implemented to test our prediction. 

\section{Conclusion}
We have demonstrated the possibilities of a self-driving car using a simple Arduino board and a mobile phone camera.
Over the duration of the project, a number of approaches have been tested: Computer Vision, Floor Detection and Imitation Learning.
The final result was a CNN algorithm which was able to take image input and output a directional command which approximated the human label set at 95\% accuracy.
The model car was able to navigate the indoor environment effectively while avoiding immediate contact with it's surroundings.
Some of the pitfalls of the algorithm stem from a lack of additional information, causing unexpected behavior in unfamiliar environments.

Due to the limited testing set used to train the car, various orientations of the car relative to its environment introduce the potential for unpredictable movement.
Implementing additional algorithms in conjunction to imitation learning is advised, as well as expanding the training set to include uniformly numerous instances of all desired behavior.

The most difficult aspect of this project was deciding on which approach would bring us closer to our desired goal of navigating the indoor environment.
We attempted to explore as many approaches as possible (outlined above) while not sinking all of our time on any one approach.
This was due to our inexperience with the domains involved in this project, which consequently gave us a very broad learning experience. 

All of the technical aspects of this projects can be found at \url{https://github.com/RoboticsCourse}

\section{Acknowledgements}
We would like to thank Lisa Zhang for the invaluable help in our attempt to build a Convolution Neural Network for floor detection.
It showed great promise and would have been explored further given more time to work on the project.
We would also like to thank Florian Shkurti for his initial idea behind imitation learning and subsequent proposals for improvements of the algorithm.
Most of all, we would like to thank Arnold Rosenbloom, our coordinator professor, for his continual support and contributions over the duration of the project.
If it were not for his generous help, this project would have never been completed in time.
Thank you, Arnold.



\section*{References}

[1]  J. Canny. “A Computational Approach for Edge Detection”. IEEE
Transactions on Pattern Analysis and Machine Intelligence, 8(6):679-698, 1986.\\ [0.1in]
[2]  Y. Li and S. Birchfield. “Image-Based Segmentation of Indoor Corridor Floors for a Mobile Robot”. 1-7, 2018.\\ [0.1in]

\end{document}